\documentclass[journal,final]{IEEEtran}%
\usepackage{array}
\usepackage[comma, numbers]{natbib}
\usepackage{xcolor}
\usepackage{amsmath}
\usepackage{amsfonts}
\usepackage{amssymb}
\usepackage{booktabs}
\usepackage{graphicx} 
\usepackage{multirow}
\usepackage{pifont}
\newcommand{\xmark}{\textit{\sffamily X}}%

\newcommand{\PARbegin}[1]{\noindent {\bf #1~}}

\begin{document}
\title{Real-time Human-Centric Segmentation for Complex Video Scenes}

\author{Ran Yu, Chenyu Tian, Weihao Xia, Xinyuan Zhao, Haoqian Wang, Yujiu Yang

\thanks{This work has been submitted to the IEEE for possible publication. Copyright may be transferred without notice, after which this version may no longer be accessible.}
\thanks{R.~Yu, C.~Tian, W.~Xia, H.~Wang, and Y.~Yang are with Tsinghua Shenzhen International Graduate School, Tsinghua University, China. (Corresponding author: Yujiu Yang)}%
\thanks{X.~Zhao is with Northwestern University, USA.}
}

\maketitle

\IEEEdisplaynontitleabstractindextext

\IEEEpeerreviewmaketitle

\begin{abstract}
Most existing video tasks related to ``human'' focus on the segmentation of salient humans, ignoring the unspecified others in the video. 
Few studies have focused on segmenting and tracking all humans in a complex video, including pedestrians and humans of other states (\textit{e.g.}, seated, riding, or occluded). 
In this paper, we propose a novel framework, abbreviated as HVISNet, that segments and tracks all presented people in given videos based on a one-stage detector. 
To better evaluate complex scenes, we offer a new benchmark called HVIS (Human Video Instance Segmentation), which comprises 1447 human instance masks in 805 high-resolution videos in diverse scenes. 
Extensive experiments show that our proposed HVISNet outperforms the state-of-the-art methods in terms of accuracy at a real-time inference speed (30 FPS), especially on complex video scenes. 
We also notice that using the center of the bounding box to distinguish different individuals severely deteriorates the segmentation accuracy, especially in heavily occluded conditions.
This common phenomenon is referred to as the ambiguous positive samples problem.
To alleviate this problem, we propose a mechanism named Inner Center Sampling to improve the accuracy of instance segmentation. 
Such a plug-and-play inner center sampling mechanism can be incorporated in any instance segmentation models based on a one-stage detector to improve the performance. 
In particular, it gains 4.1 mAP improvement on the state-of-the-art method in the case of occluded humans.
Code and data are available at \texttt{https://github.com/IIGROUP/HVISNet}.
\end{abstract}

\IEEEpeerreviewmaketitle

\section{Introduction}
\label{sec:intro}

\begin{figure*}[th]
\centering
\includegraphics[width=\textwidth]{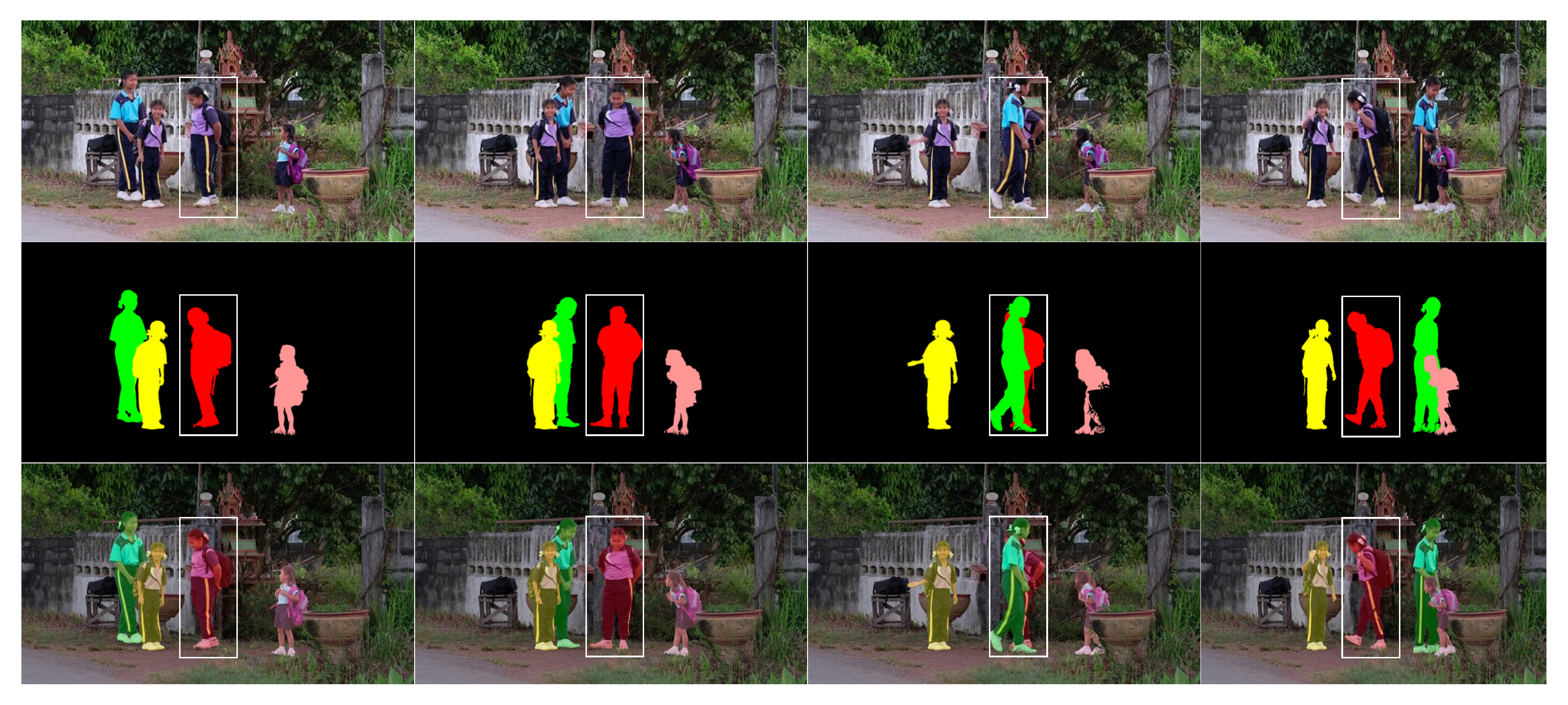}
\caption{A case visualization of human-centric video segmentation under complex real-world scenarios. The three rows are input video, output result, and the two superimposed visualization effects, respectively. 
The mask color of the output result indicates the identity of the humans. 
The little girl with the white bounding box has undergone complex scenes such as occluded, overlapped, and reappeared.}
\label{fig:teaser}
\end{figure*}

\IEEEPARstart{H}uman-centric research on tracking, detection, and segmentation has gained significantly increased interest due to its broad application scenarios, such as autonomous driving, intelligent surveillance, human-machine interaction, and mobile entertainment. 
Such research in the image domain, such as portrait segmentation~\cite{shen2017high,chen2019boundary,shen2016automatic}, person re-identification~\cite{zeng2020illumination,ye2016person,zheng2015scalable}, or pedestrian detection~\cite{qian2019oriented,zhang2018occluded,zhang2020attribute,zhang2020stinet,huang2020nms}, has been well-studied and applied in real-life applications. 
Its counterpart in the video, however, is an issue that has not been sufficiently addressed.
Current studies on video instance segmentation are primarily concentrated on simple multi-category scenes without distinguishing objects of the same categories, \textit{e.g.}, pedestrians in the crowd. Meanwhile, the capability of handling heavily-occluded people is an urgent requirement for practical applications such as autonomous driving.
Therefore, in this paper, we propose a novel framework that extends HIIS from the image domain to the video domain, focusing on instance segmentation of multi-humans under complex real-world scenarios. 
To be specific, our goal is to accurately segment every human in given videos and guarantee a consistent identity for the same person, despite reappearance after several frames. 

This \textit{human-centric video instance segmentation for complex scenes} (HVIS-CS) is more challenging than some related tasks such as human image instance segmentation (HIIS) \cite{zhang2019pose2seg,bolya2019yolact,lee2020centermask}, multi-object tracking and segmentation (MOTS) \cite{voigtlaender2019mots}, and video instance segmentation (VIS) \cite{perazzi2016benchmark}. 
Compared to HIIS, it requires instance segmentation on each frame of the video and needs to ensure that the identity consistency of each human, as illustrated in Figure~\ref{fig:teaser}.
MOTS extends the basic task of multi-object tracking to the pixel level and uses a more accurate mask to represent the object. Different from MOTS, HVIS-CS segments all humans accurately, whether this human is a pedestrian. MOTS only segments and tracks primary pedestrians in the video, ignoring riders, sitting and standing persons. 
The VIS task extends image instance segmentation to the video, aiming to simultaneously detect, segment, and track \textit{object} instances in videos. 
Our HVIS-CS can be taken as a sub-task of the VIS, focusing on distinguishing different \textit{human} instances in complex video scenes.
It requires identifying every shown person and ensuring consistency of inter-frame identity under complex scenarios such as overlapping, occlusion, disappearance, and reappearance. 

\begin{figure*}[th]
  \centering
  \includegraphics[width=0.9\textwidth]{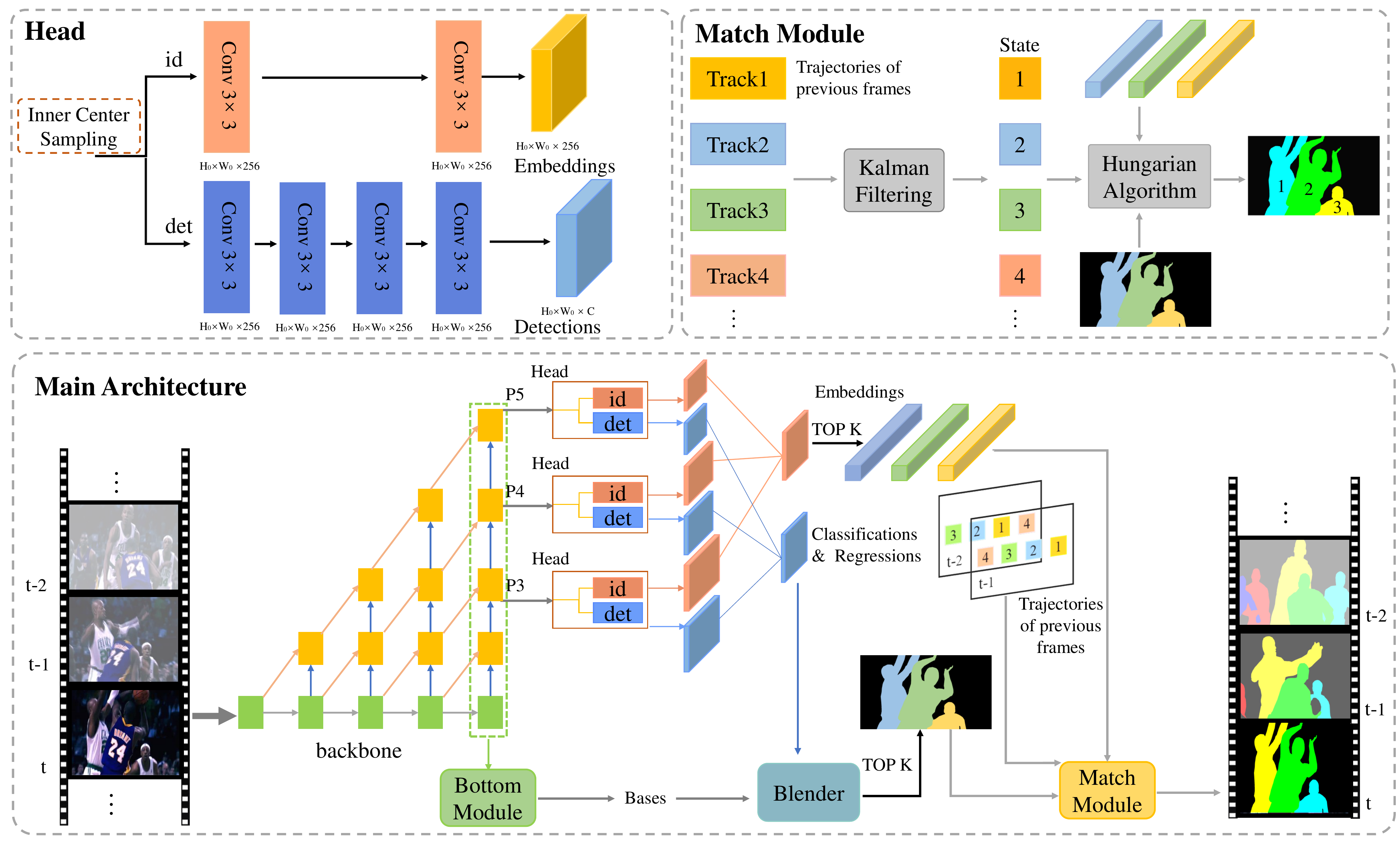}
  \caption{
  The pipeline of the proposed method. Given the current t-frame in a video sequence, we obtain the multi-level features $P_3$, $P_4$, $P_5$, which is represented as $H_0 \times W_0 \times C_0$, $\frac{1}{2} H_0 \times \frac{1}{2} W_0 \times C_0 $, $\frac{1}{4} H_0 \times \frac{1}{4}W_0 \times C_0$, respectively.
  These features, taking $P_3$ ($H_0 \times W_0$) as an example, are then fed into the head structure composed of two branches to extract the detected targets $(H_0 \times W_0 \times C)$ and the appearance embedding $(H_0 \times W_0 \times 256)$. 
  The detection branch contains two components to estimate classification score $(H_0\times W_0\times 1)$ and regression $(H_0 \times W_0 \times 4)$, respectively. Then we select top $K$ detections and embeddings for identification matching based on the classification scores from the multi-level heads. Finally, we use the embeddings and bounding boxes of the current frame to associate tracks in the previous frame to obtain the identity of the human and update the tracks.
  }
  \vspace{-10pt}
  \label{fig:framework}
\end{figure*}

Despite good application prospects, the research on HVIS-CS is still relatively scarce. 
Currently, existing VIS studies mainly use MaskRCNN~\cite{he2017mask} as the basic framework, which first obtains the bounding boxes through MaskRCNN and then extracts the feature of the bounding boxes to complete the trajectory matching. 
These methods are equivalent to using two separate models to locate the bounding box's position on the image and extract features for each bounding box, then match the bounding box with the existing trajectories based on these features in a video. However, using two networks to extract features separately is not conducive to the real-time performance of video instance segmentation. Therefore, we propose a novel framework called HVISNet for marking humans accurately in videos based on BlendMask~\cite{chen2020blendmask}, a state-of-the-art method for HIIS. 
We add a new head branch, which is parallel with the detection branch, aiming to extract the appearance embedding of each human instance through the backbone network. 
The framework can simultaneously obtain embedding for detection and appearance, which contributes to improving inference speed. 

One-stage detectors generally make dense predictions on the features obtained through the backbone. 
It classifies the features into positive and negative samples according to specific rules. 
Positive samples are then classified and regressed. 
For the classification of positive and negative samples, FCOS~\cite{tian2019fcos} maps the points on the feature map to the original image, and it is considered a positive sample if it falls within ground-truth. 
FCOSv2 decreases the positive sample region to within $\frac{1}{2}$ stride of the ground-truth (bounding box) center, and the performance is significantly improved. Thus, positive and negative samples have a significant impact on performance. 
Suppose we use rules in FCOS to classify positive and negative samples in intense overlapping scenes. 
In that case, the experiment finds that some positive samples belong to both human A and human B. 
These positive samples are used not only for the acquisition of the human A mask and appearance embedding but also for human B. 
Such positive samples are ambiguous. 
Therefore, to alleviate the problem of positive sample ambiguity, we propose inner center sampling to replace the original positive and negative sample classifier to distinguish positive samples of different instances in intense overlapping scenarios. 
We verify the generalization ability of this plug-and-play mechanism by incorporated it into different instance segmentation methods.

Existing datasets for similar tasks either do not exhaustively label all people or only contain simple scenes in the video. 
A human object is a particular category in the field of computer vision with many application scenarios. 
Autonomous driving, human monitoring, mobile entertainment applications, and other practical requirements need to segment and track all humans in the video, especially in complex video scenes. 
In real life, there are always severe occlusions or shifts between human bodies. 
Therefore, we propose a new benchmark called \textsc{\textbf{H}uman \textbf{V}ideo \textbf{I}nstance \textbf{S}egmentation} (HVIS), which focuses on complex real-world scenarios with sufficient human instance masks and identities.
Our dataset contains 805 videos with 1447 detailedly annotated human instances.
It also includes various overlapping scenes, which integrates into the most challenging video dataset related to humans.
The contributions of our work are three-fold:
\begin{itemize}
\item We propose a novel framework (HVISNet) for human video instance segmentation based on a one-stage detector, which outperforms the state-of-the-art methods in terms of accuracy and runs over 30fps in inference.
\item We propose a new benchmark named HVIS that focuses on complex video scenes with sufficient human instance masks and identities.
\item We propose an inner center sampling mechanism to effectively alleviates the problem of positive sample ambiguity in instance segmentation. Besides, such a plug-and-play inner center sampling mechanism shows good generalization ability and can be incorporated in any instance segmentation model based on a one-stage detector to improve human instance segmentation accuracy.
\end{itemize}

\section{Related Work}
\label{sec:related_work}

\PARbegin{HIIS.}
The typical work of human-centric image instance segmentation is combined with the human pose estimation \cite{papandreou2017towards,papandreou2018personlab}. 
For example, Pose2Instance \cite{tripathi2017pose2instance} proposes a cascaded network to apply human pose estimation for instance segmentation. Pose2seg~\cite{zhang2019pose2seg} proposes an Affine-Align operation for selecting ROIs based on pose instead of bounding-boxes. It concatenates the human pose skeleton feature to the image feature in the network to further improve the performance. 
The aforementioned methods depend on pose estimation performance, and the speed is much lower than general instance segmentation. 
The general image instance segmentation methods~\cite{tian2020conditional,xie2020polarmask} based on a one-stage detector also have good performance in the category of the human on the COCO dataset~\cite{lin2014microsoft}. 
Similar to HIIS, HVIS-CS needs to segment the human instance in every frame but also requires consistent identities for different persons across frames. 

\PARbegin{MOTS.} 
MOTS performs multi-target tracking and segmentation simultaneously.
TrackR-CNN~\cite{voigtlaender2019mots} adds an association head generating correlation vectors on the MaskRCNN~\cite{he2017mask} and uses the video timing information through 3D convolution.
PointTrack~\cite{xu2020segment,xu2020pointtrack++} proposes an efficient segmentation-based instance embedding method, which generates a novel point-by-point tracking paradigm by converting the compact image. The MOTS task only tracks the cars and the pedestrians in video and treats the cyclists, riders, and standing humans as the background without processing. It is inconsistent with practical applications.
HVIS-CS corrects this problem of MOTS by segmenting and tracking anyone who appears in the video.

\PARbegin{VIS.}
VIS is a newly proposed task that enables simultaneous detection, segmentation, and tracking object instances in videos. 
For example, MaskTrack~\cite{yang2019video} adds a tracking branch to MaskRCNN, used to assign the identity of each instance. 
Lin~\textit{et.al.}~\cite{lin2020video} propose a modified VAE built on top of MaskRCNN for instance-level video segmentation and tracking.
STEm-Seg~\cite{athar2020stem} is a bottom-up method for clustering each instance pixel and introduce timing information through 3D convolution.
These methods aim to segment forty categories of objects into video instances, and the scenes are relatively simple. 
When objects of different categories are matched between frames, the tracking is completed using classified features, but the inter-frame matching between different instances of the same category is ignored. 
Differently, our HVIS-CS task focuses on how to represent each instance in complex video scenes with a high-quality pixel mask.

\begin{figure*}[!ht]
  \centering
  \vspace{-2mm}
  \begin{tabular}{ccccc}
    \hspace{-3.2mm}
    \includegraphics[width=0.195\linewidth]{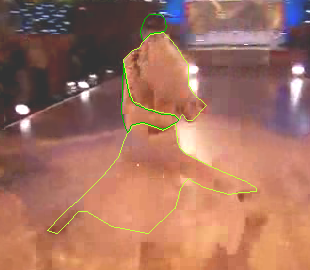}   
    & \hspace{-6mm}
    \includegraphics[width=0.195\linewidth]{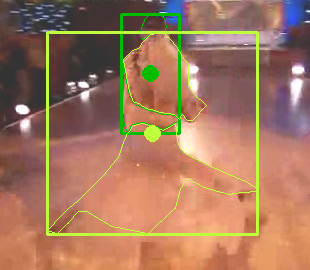}  & \hspace{-6mm}
    \includegraphics[width=0.195\linewidth]{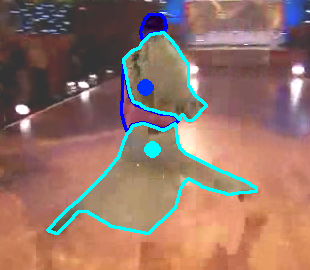}  & \hspace{-6mm}
    \includegraphics[width=0.195\linewidth]{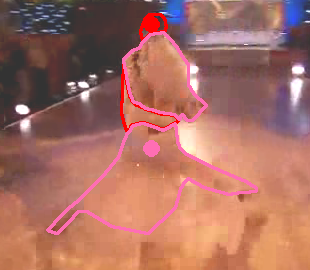}  & \hspace{-6mm}    
    \includegraphics[width=0.195\linewidth]{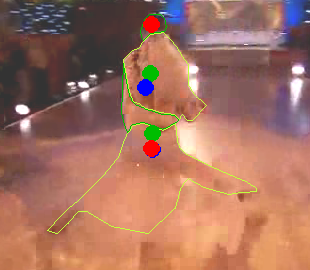} \\
    \hspace{-3.2mm}
    \includegraphics[width=0.195\linewidth]{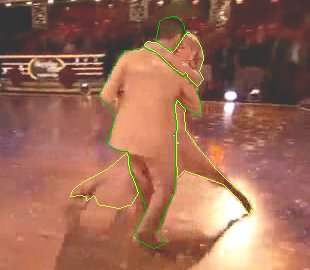}   & \hspace{-6mm}
    \includegraphics[width=0.195\linewidth]{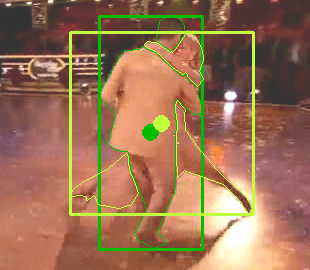}  & \hspace{-6mm}
    \includegraphics[width=0.195\linewidth]{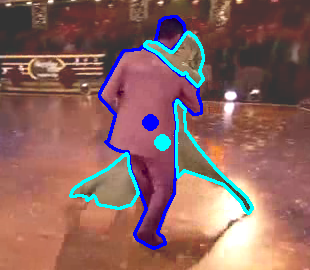}  & \hspace{-6mm}
    \includegraphics[width=0.195\linewidth]{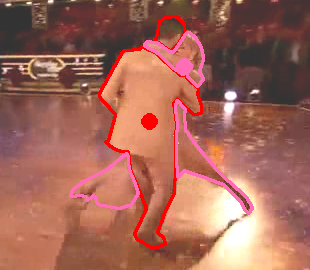}  & \hspace{-6mm}
    \includegraphics[width=0.195\linewidth]{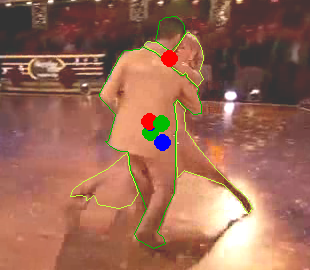} \\
    (a) Original image & (b) Center of bbox & (c) Centroid of mask & (d) Inner center & (e) Center points
    
  \end{tabular}
  \vspace{-2mm}
 VIS is a newly proposed task that enables \caption{Comparison of different sampling centers. The two rows represent different frames in a video, with five columns representing the original image (a), center of the bounding box (b), centroid of the mask (c), our proposed inner center (d), and representation of all centers (e). The green points denote the center of the bounding box, blue points denote the centroid of the mask, and red points denote the inner center.}
  \vspace{-12pt}
  \label{fig:point}
\end{figure*}

\section{HVISNet Framework}
\label{sec:method}
Our goal is to mark every human in complex video scenes accurately. 
To be specific, given a video sequence $F=\left\{ f_1,f_2,\ldots,f_i,...,f_n \right\}$, we obtain the accurate mask and appearance feature in each frame image $f_i$, then we associate the appearance feature with the previous frame $\{f_k\}_{k=1}^{i-1}$ to achieve the human identification in the video. 

The proposed overall framework named HVISNet is shown in Figure~\ref{fig:framework}, which is built upon one-stage detector FCOS~\cite{tian2019fcos}. 
We add an identification branch upon the backbone network, which is parallel to the detection branch. 
The identification branch is used for extracting the appearance embedding of each instance. 
As for the segmentation task, we follow the \textit{top-down meets bottom-up} structure proposed in BlendMask~\cite{chen2020blendmask} and retain the bottom module and blender. 
For the problem of ambiguous positive samples in the training process, we have explored the impact of positive samples on instance segmentation and propose the inner center sampling strategy to alleviate the ambiguous positive samples problem. 
Furthermore, we construct the match module to associate the appearance embedding obtained from different frames in a video. More details are covered in the following sections.

\subsection{Identification Branch}
\label{subsec:method_identification}

As illustrated in Figure~\ref{fig:framework}, the multi-level backbone feature maps are extracted from each frame image with a resolution of $H \times W $, and represented as $P_3$ ($H_0 \times W_0 \times C_0$), $P_4$ ($\frac{1}{2} H_0 \times \frac{1}{2} W_0 \times C_0 $), and $P_5$ ($\frac{1}{4} H_0 \times \frac{1}{4}W_0 \times C_0 $) respectively, where $C_0$ is the number of channels. 
And the multi-level backbone feature maps are fed into the head structure composed of two branches, \emph{i.e.} the identification branch (marked as id) and the detection branch (marked as det) to extract the appearance embedding and the detected targets, respectively. 

The identification branch aims to generate the appearance embedding to associate the same human and distinguish different humans in a video. And it is composed of two convolution layers to extract the appearance embedding from each position on the multi-level feature maps. 
The appearance embedding is expressed as $E_{x,y}=\{E_{(x_1,y_1)},E_{(x_2,y_2)},\cdots,E_{(x_n,y_n)}\} \in \mathbb{R} ^{1 \times 1 \times 256}$, where $(x, y)$ indicates the position on the feature map and $(x_i,y_i)$ represents the position of each positive sample on the feature map, both here and below.
For each feature map, we apply the sampling strategy proposed in FCOS \cite{tian2019fcos} to determine the positive samples $P=\{p_{(x_1,y_1)},p_{(x_2,y_2)},...,p_{(x_n,y_n)}\}$.
These positive samples $P$ are then fed into different heads module to predict the categories, offsets, and appearance embedding.
Obviously, the distance between the appearance embedding from different humans should be greater than that from the same human. 
We take those appearance embedding with identical labels as the positive samples and those with different labels as the negative ones.
Thus, the appearance embedding can be learned with the triplet loss~\cite{schroff2015facenet}:
\begin{equation}
	 L_{tri} =
	 \max(||E_{x,y}^a - E_{x,y}^p||_2^2 - ||E_{x,y}^a - E_{x,y}^n||_2^2 + \alpha, 0),
\end{equation}
where $\alpha$ is the bound margin, $E_{x,y}^a$ is the anchor, $E_{x,y}^p$ is the positive sample, and $E_{x,y}^n$ is the negative sample.

To alleviate the problem of easily falling into the local optimum with only the triplet loss, we introduce the classification loss to assist the training:
\begin{equation}
    L_{cls} = \mathbb{CE}(\phi(E_{x,y}), l_{x, y}),
\end{equation}
where $\phi$ is a simple binary classifier, $l_{x,y}$ is the label of $E_{x, y}$, and $\mathbb{CE}$ denotes the cross-entropy.
We set the label of negative samples to 0 and the label of positive samples to the ground-truth human id.
In a conclusion, the identification branch is trained with the triple loss $L_{tri}$ and the classification loss $L_{cls}$:
\begin{equation}
L_{id} = L_{tri}+L_{cls}.
\end{equation}

\subsection{Inner Center Sampling}
\label{subsec:method_sampling}

We design HVISNet directly according to the previous section but find that the segmentation accuracy and appearance embedding differentiation in complex scenes is poor. By analyzing these underperforming cases, we find that there exist some ambiguous positive samples. We use an example to illustrate what are ambiguous positive samples. Note that the two humans are represented as A and B as shown in Figure~\ref{fig:point} (a). The classification of the positive and negative samples of the two humans is based on the position of the center of the bounding box. That is, if the sample position within $\frac{1}{2}$ stride range around the center, the sample is classified as a positive sample. However, the bounding box centers of the two humans A and B in the figure are close, which means the positive sample area overlaps seriously. Moreover, the two humans are similar in size and cannot be distinguished by the different levels. These overlapping positive samples need to represent both the embedding of A and the embedding of B. During training, there is a problem that the same positive sample has multiple labels. We define such a problem as the ambiguity positive samples problem.

\begin{figure}[!th]
  \centering
  \small
  \vspace{-2mm}
  \begin{tabular}{ccc}
    \hspace{-3.2mm}
    \includegraphics[width=0.325\linewidth]{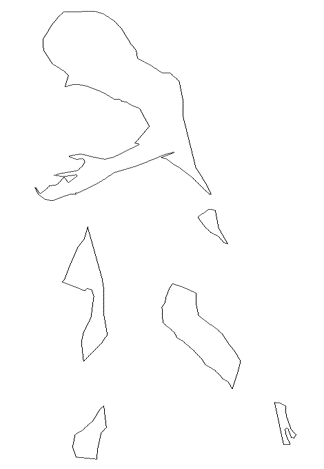}   & \hspace{-6mm}
    \includegraphics[width=0.325\linewidth]{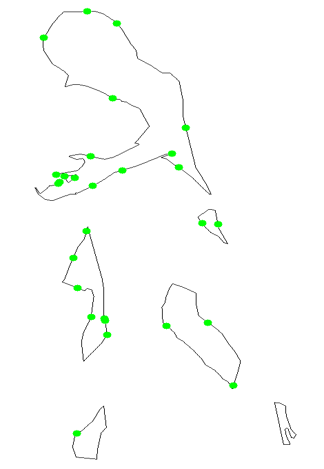}  & \hspace{-6mm}
    \includegraphics[width=0.325\linewidth]{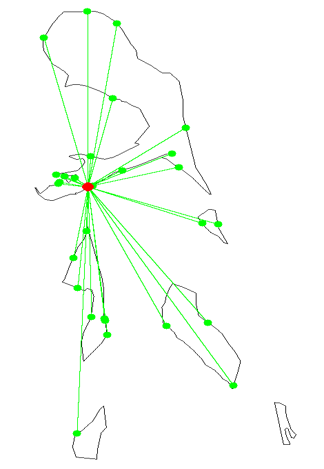} \\
    (a) & (b) & (c) 
  \end{tabular}
  \vspace{-2mm}
  \caption{Definition of the Inner Center. The three columns represent the three steps of finding the center inner, including obtaining the edge point set of the mask, sampling, and finding the point corresponding to the shortest distance.}\label{fig:inner}
  \vspace{-7pt}
\end{figure}

We find that if the center of the bounding box is used to classify positive and negative samples, then there will generate ambiguous positive samples in complex scenes. Furthermore, it is found that the positions of features with poor performance are almost outside the human mask but inside the bounding box, denote $P_{NP}$. Ideally, if the position of the positive sample falls inside the mask, it must be inside the bounding box. Therefore, if the positive samples in complex scenes are guaranteed to be inside each independent instance, we can avoid overlapping the positive samples of different instances. We also alleviate the problem of ambiguous positive samples. Thus, we propose the inner center sampling strategy, which is sampling at the \textbf{Inner Center}. Inner Center is defined as the point $(x_{in},y_{in})$ inside a human's mask and matching $\min \sum_{(x_i,y_i)\in P_E} [(x_{in}-x_i)^2+(y_{in}-y_i)^2] $, where $P_E$ is the edge sampling point set of a human mask. As shown in~\ref{fig:inner}, we first find the edge point sets of a human mask and sample edge points randomly, which are marked green. Then we find the point with the shortest distance between the sampled edge point set inside the mask.

We visualize different positions of center points in Figure~\ref{fig:point}. Column (b) is the center of the bounding box introduced above, which is easy to generate ambiguous positive samples. Column (c) is the centroid of the mask. Intuitively, the mask is more refined than the bounding box. Sampling with the centroid of the mask leads to better performance than the center of the bounding box. However, neither the center of the green bounding box nor the centroid of the blue mask can be guaranteed to be inside the human all the time. In column (d), the Inner Center marked in red is still guaranteed to be inside the human if the overlap exceeds 0.7. Column (e) contains the different center points, from which we discover that our Inner Center can alleviate the ambiguous positive samples problem and better distinguish different individuals.

\begin{table}[th]
\caption{Comparison with existing video segmentation datasets.}
\setlength\tabcolsep{10pt}
\begin{center}
\begin{tabular}{lccc}
\toprule[1.5pt]
Dataset &Exhaustion &Consistency &Complexity \\
\hline
DAVIS~\cite{pont20172017}
&&$ \checkmark$ & \\
DAVSOD~\cite{fan2019shifting}
&&&$ \checkmark$\\
MOTS~\cite{Voigtlaender19CVPR_MOTS}
&&$ \checkmark$&\\
VIS~\cite{yang2019video}
&$ \checkmark$&$ \checkmark$&\\
\hline
Our HVIS&$ \checkmark$&$ \checkmark$&$ \checkmark$\\
\bottomrule[1.5pt]
\end{tabular}
\end{center}
\label{tab:datadif}
\end{table}

\begin{table}[th]
\small
\caption{Comparison between VISPersons~\cite{yang2019video} and our proposed HVIS.}
\setlength\tabcolsep{4pt}
\begin{center}
\begin{tabular}{c|ccccc}
\toprule[1.5pt]
Dataset& \#videos & \#small-taget & \#instances  & \#overlapping \\
VISPersons &    59&  11.12 $\%$ &  77& 10  \\
HVIS &    81&49.74 $\%$  &  205  &97  \\
\bottomrule[1.5pt]
\end{tabular}
\end{center}
\label{tab:dataset_compare}
\end{table}

\begin{table}[th]
\centering
\caption{Quantitative evaluation with state-of-the-art methods on two datasets, VISPersons and HVIS.}
\setlength\tabcolsep{3pt}
\begin{tabular}{l|l|cccc}
\toprule[1.5pt]
Dataset  & Method  & sMOTA & MOTSA & MOTSP & FPS \\ \hline
\multirow{3}{*}{VISPerson} &MaskTrack & 53.1 &66.7 &83.8 &25     \\
& STEm-Seg & \textbf{56.2} &\textbf{65.7} &\textbf{86.1} &7 \\
& HVISNet &55.7 & 65.3 &86.0&\textbf{30} \\ \hline
\multirow{3}{*}{HVIS} &MaskTrack &37.3 &54.3 &78.5 &25     \\
&STEm-Seg &44.5 &62.0 &78.8 &7 \\
&HVISNet &\textbf{52.1} &\textbf{64.4} &\textbf{81.9} &\textbf{30}  \\   
\bottomrule[1.5pt]
\end{tabular}
\label{tab:result}
\end{table}

\subsection{Match Module}
\label{subsec:method_match}

The match module is to associate each frame in the video to obtain the identification of the human. Given a test video, we input each frame of the video into the network chronologically. The backbone network extracts multi-level features from each frame and inputs them into heads composed of a detection branch and identification branch. The outputs of detection are category score and regression of bounding box. And appearance embedding is obtained from the identification branch. For the first frame, we initialize the track according to the bounding boxes, which means we number each person detected on the first frame. On subsequent frames, appearance embedding and bounding boxes are inputted to the match module to associate with the predictions of the previous frame, which means the humans in subsequent frames are numbered according to the results of the previous frame. The match module is equivalent to completing the post-processing steps of the video frame through Kalman filtering ~\cite{Maybeck90Kalman} and Hungarian matching for each frame of the image.

\section{HVIS Benchmark}
\label{sec:data}

To evaluate the performance of our method in complex video scenes, we propose the HVIS benchmark.

\subsection{Dataset Composition}
\label{subsec:data_composition}

The proposed HVIS benchmark, including 1447 human instances in 805 videos, is divided into a training set and a validation set. 
The proposed dataset has the following characteristics. 
\textbf{Exhaustion}: all human instances that ensure that each video should be annotated at the pixel level regardless of the size and pose.
\textbf{Consistency}: the identity of each human instance in the video should be unchanged. 
\textbf{Complexity}: the diversity of video scenes should be guaranteed, which means the dataset needs to contain various complex cases, such as occlusion, fast motion, strange poses, and overlapping, \emph{etc.}
We obtain this dataset by reprocessing existing video segmentation datasets of related tasks~\cite{brox2010object,ochs2013segmentation,galasso2013unified,sundberg2011occlusion, Voigtlaender19CVPR_MOTS, fan2019shifting, yang2019video, pont20172017, MOT16} rather than collecting and annotating from scratch.
We select video data from the datasets above and do cropping, relabeling, and other operations to build our dataset.
Specifically, DAVIS~\cite{pont20172017} is the dataset of the video object segmentation task, which refers to segmenting and tracking the specified human in the first frame. 
That is to say, the non-specified human or the human that appears in the subsequent frames is not labeled. 
DAVSOD~\cite{fan2019shifting} is proposed for the task of video saliency segmentation salient humans in the video and ignore the non-salient human.
MOTS~\cite{Voigtlaender19CVPR_MOTS} is to segment and track pedestrians, which does not label humans in other states such as riding, sitting, or standing. 
Thus, the above datasets are not fully labeled. 
VIS~\cite{yang2019video} is proposed for the video instance segmentation task, which is marked all humans in the video. But it was found that there were a certain number of missing labels. In terms of continuity, video saliency segmentation only needs to segment the saliency target, which is not needed to consider the target's identity in the video. Other video segmentation tasks need to consider the identity, and they meet the continuity. The complexity is illustrated in detail in the next section. 
The comparison with existing video segmentation datasets is shown in Table~\ref{tab:datadif}.

\begin{figure*}[!th]
  \centering
  \vspace{-5mm}
  \begin{tabular}{m{2mm}ccccc}
    \hspace{-3.2mm}
    {\vspace{-1.2cm}}{(a)} & \ & \  & \  & \  & \\
    \specialrule{0mm}{-0.3mm}{-0.3mm}
    \hspace{-3.2mm}
    {\vspace{-1.2cm}}{(b)} & \hspace{-5mm}
    \includegraphics[width=0.185\linewidth]{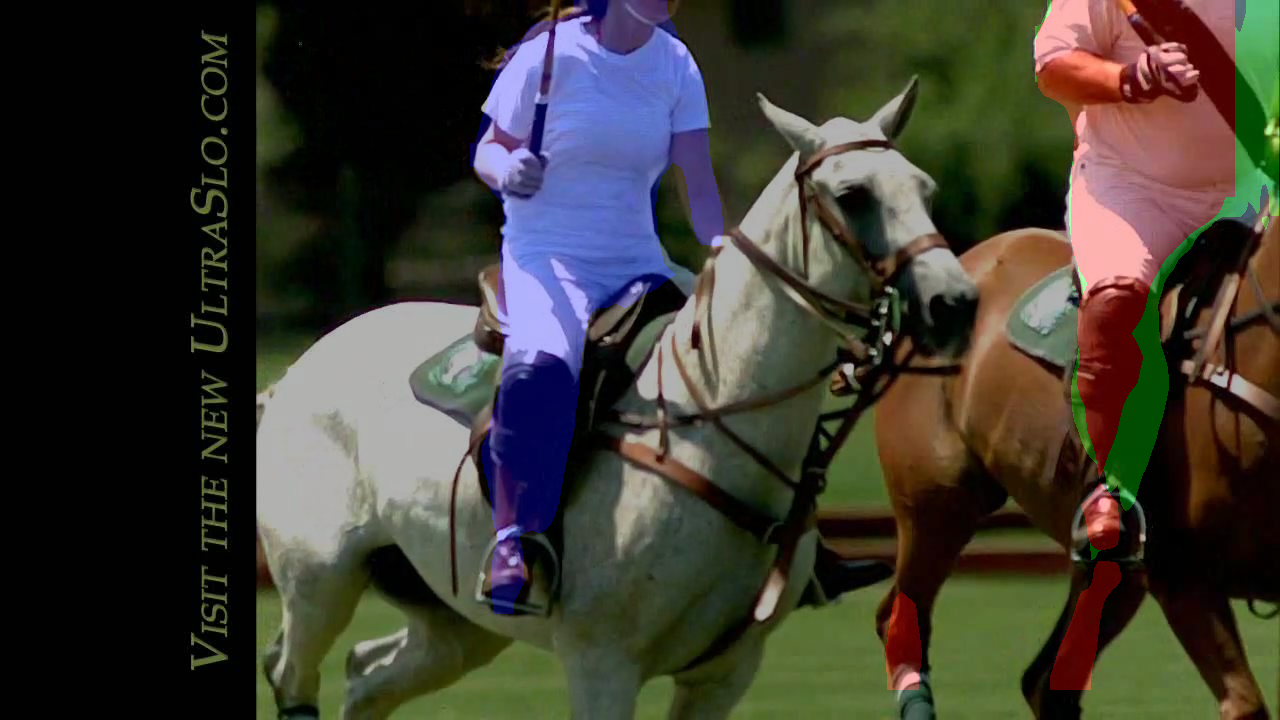}  & \hspace{-5mm}
    \includegraphics[width=0.185\linewidth]{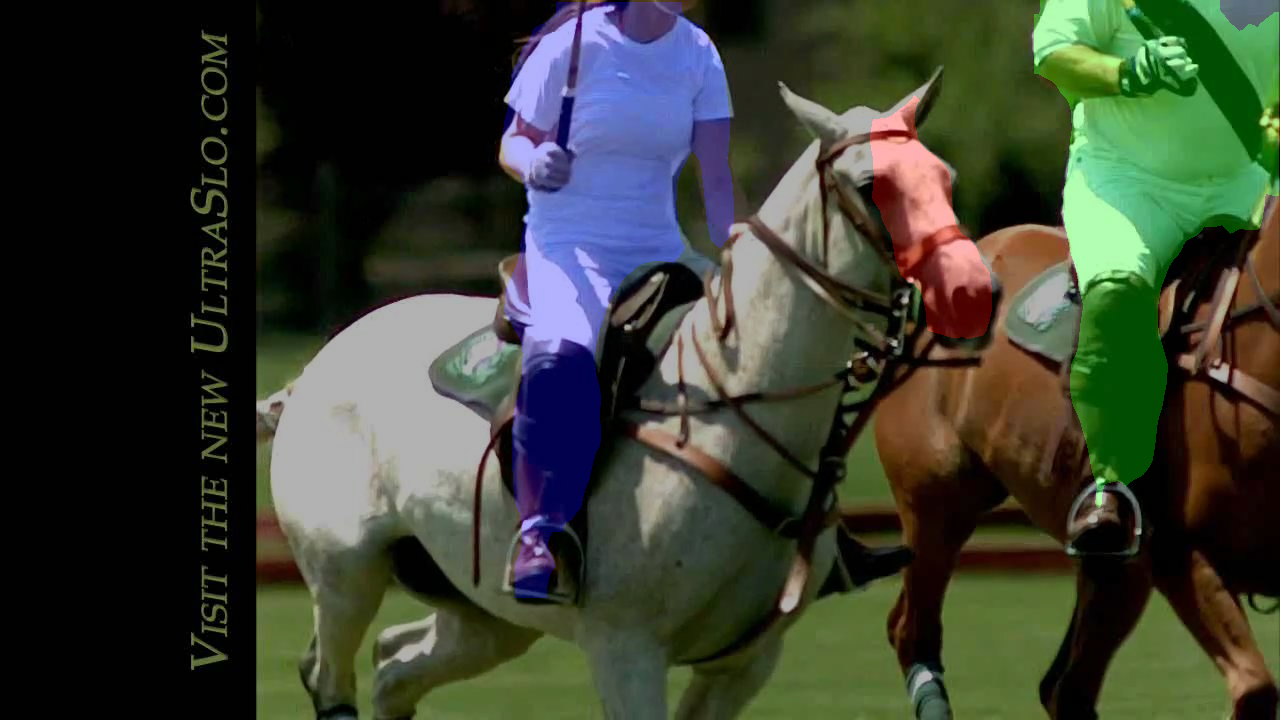}  & \hspace{-5mm}
    \includegraphics[width=0.185\linewidth]{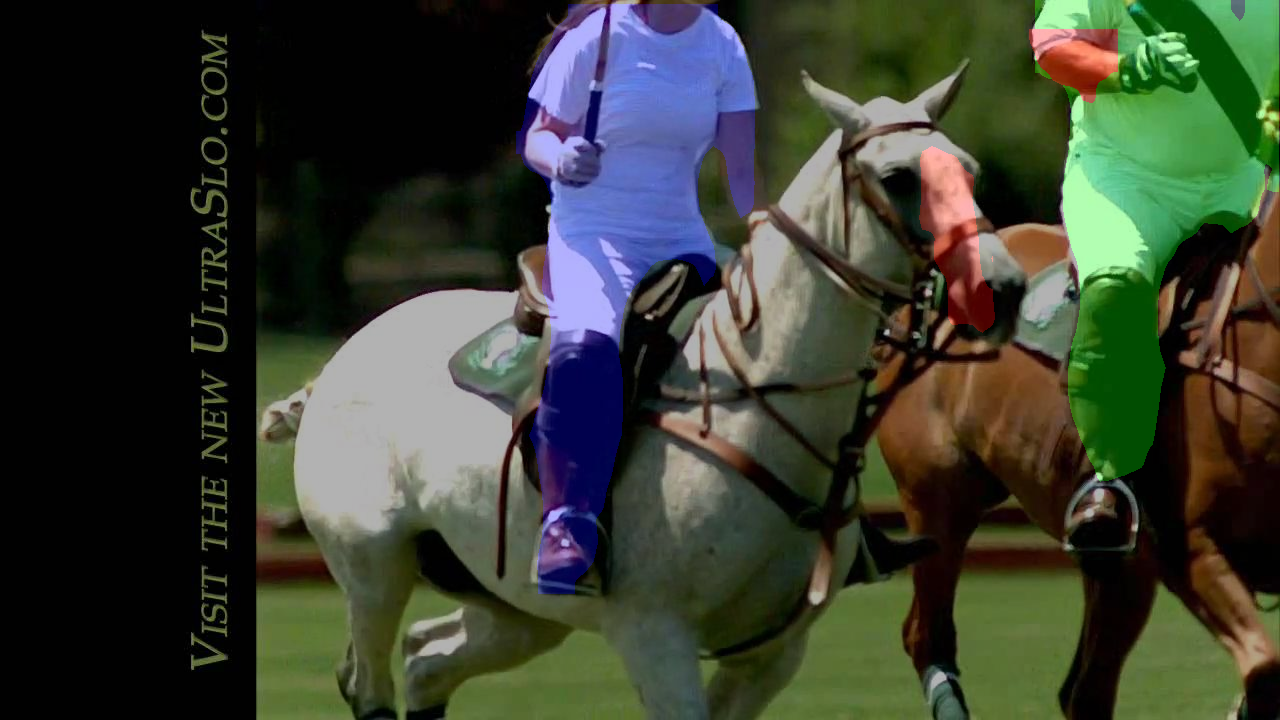}  & \hspace{-5mm}
    \includegraphics[width=0.185\linewidth]{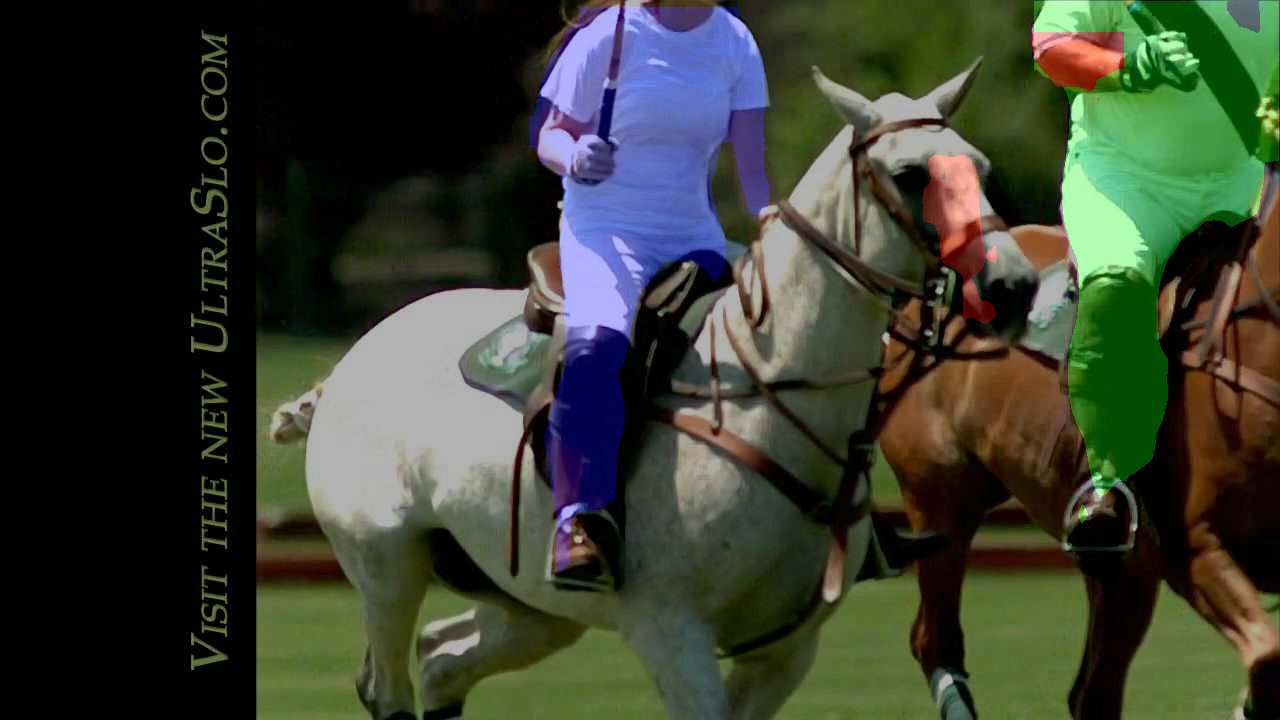}  & \hspace{-5mm}    
    \includegraphics[width=0.185\linewidth]{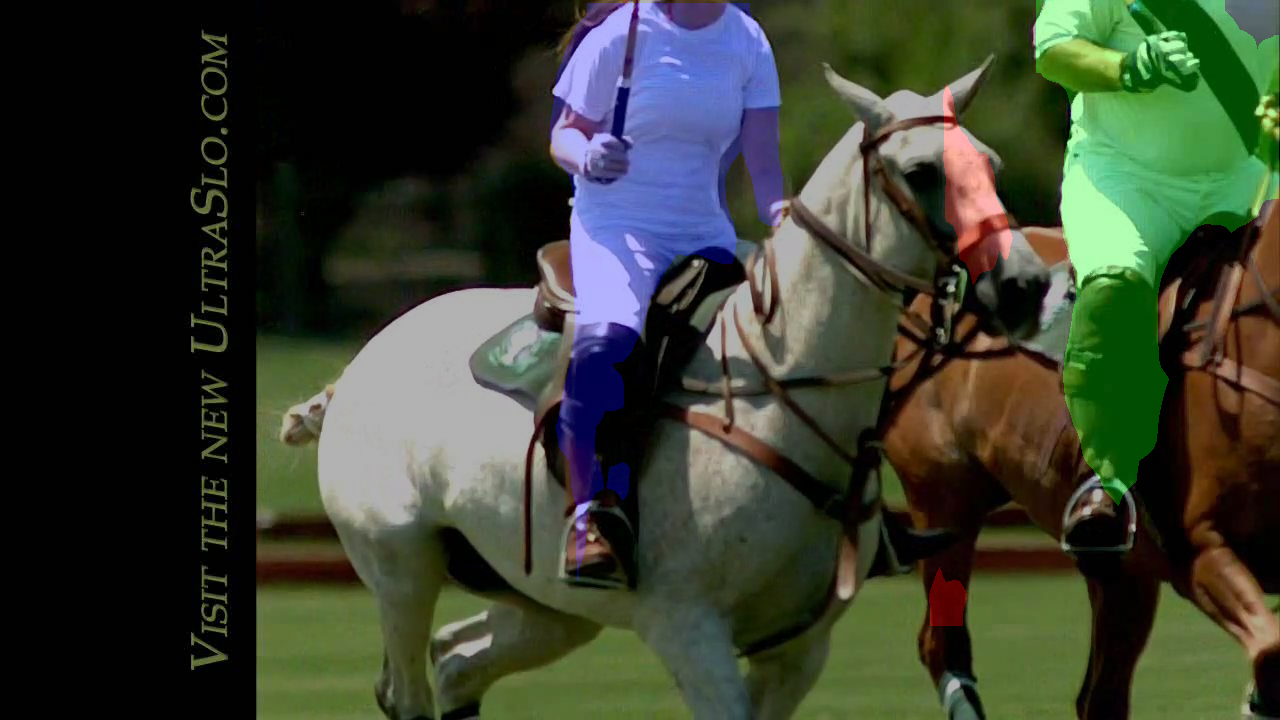} \\
    \specialrule{0mm}{-0.3mm}{-0.3mm}
    \hspace{-3.2mm}
    {\vspace{-1.8cm}}{(c)} & \hspace{-5mm}
    \includegraphics[width=0.185\linewidth]{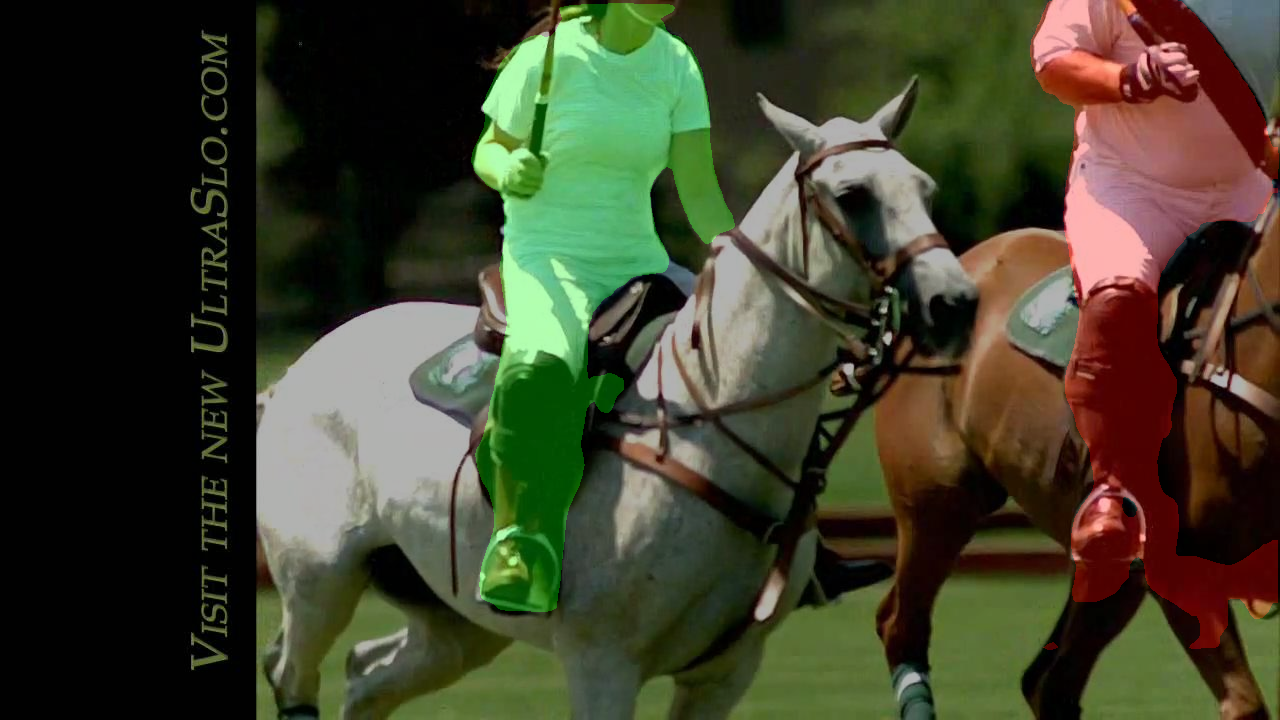}   & \hspace{-5mm}
    \includegraphics[width=0.185\linewidth]{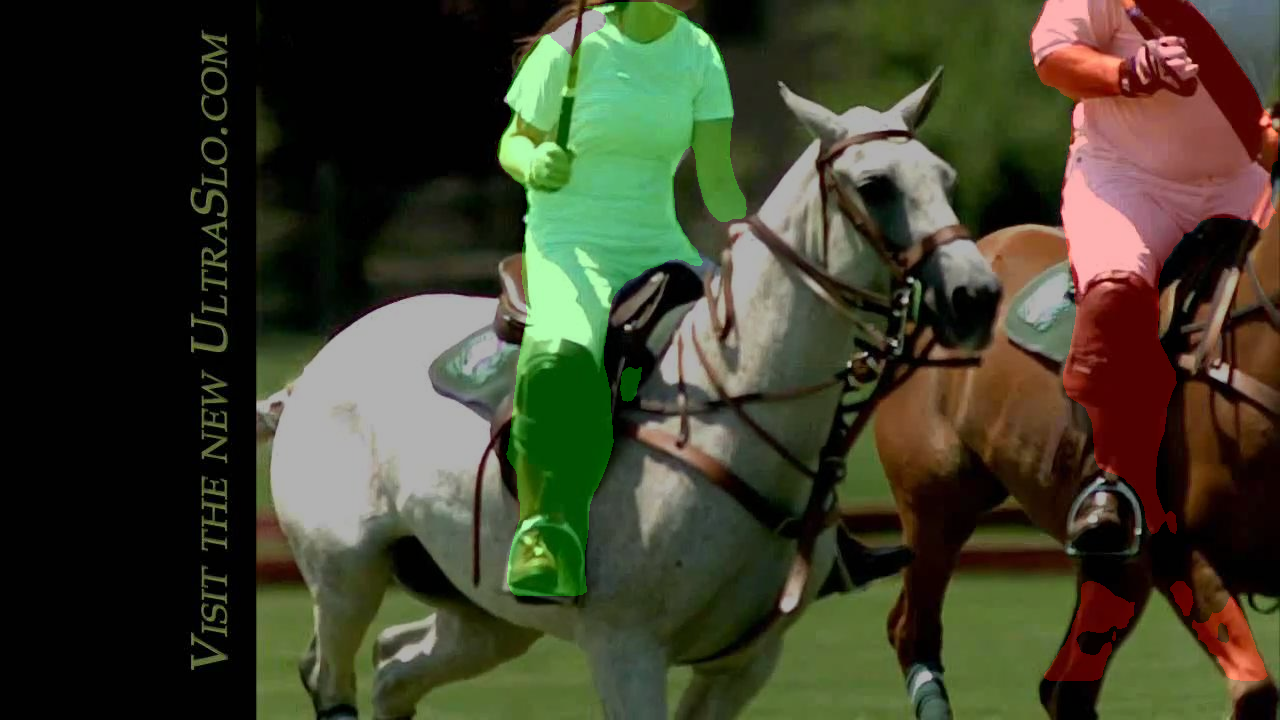}  & \hspace{-5mm}
    \includegraphics[width=0.185\linewidth]{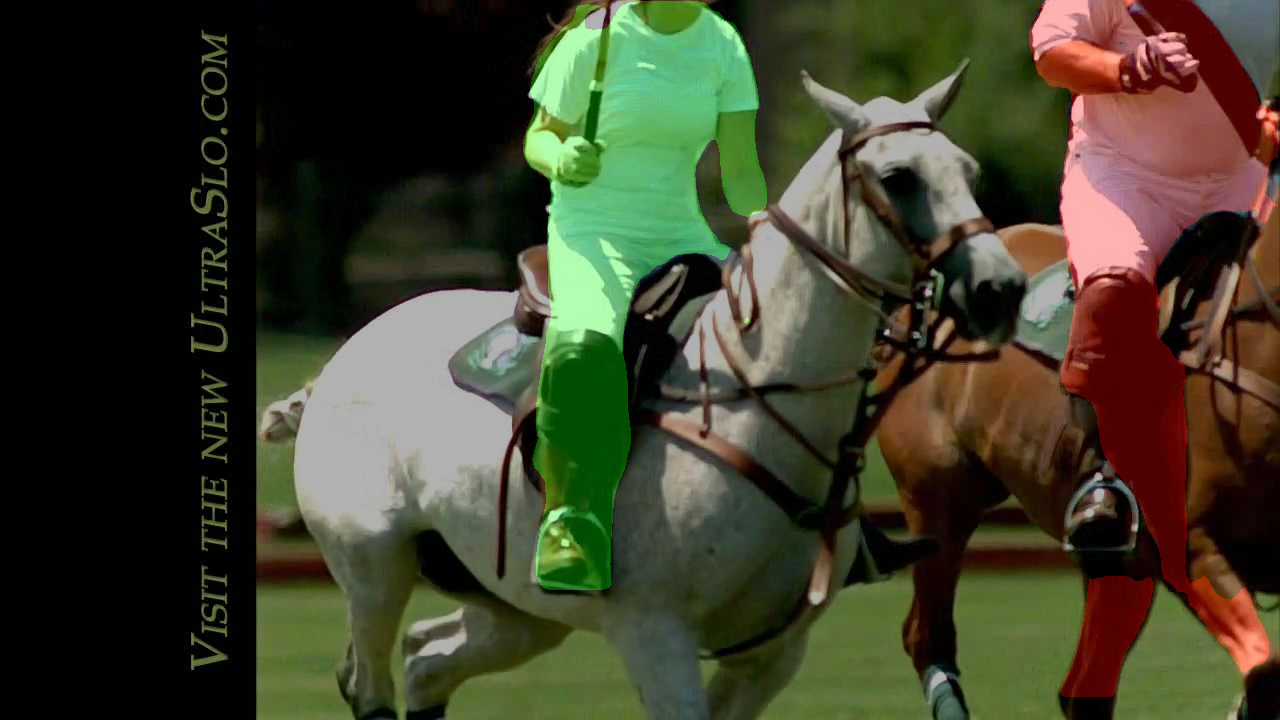}  & \hspace{-5mm}
    \includegraphics[width=0.185\linewidth]{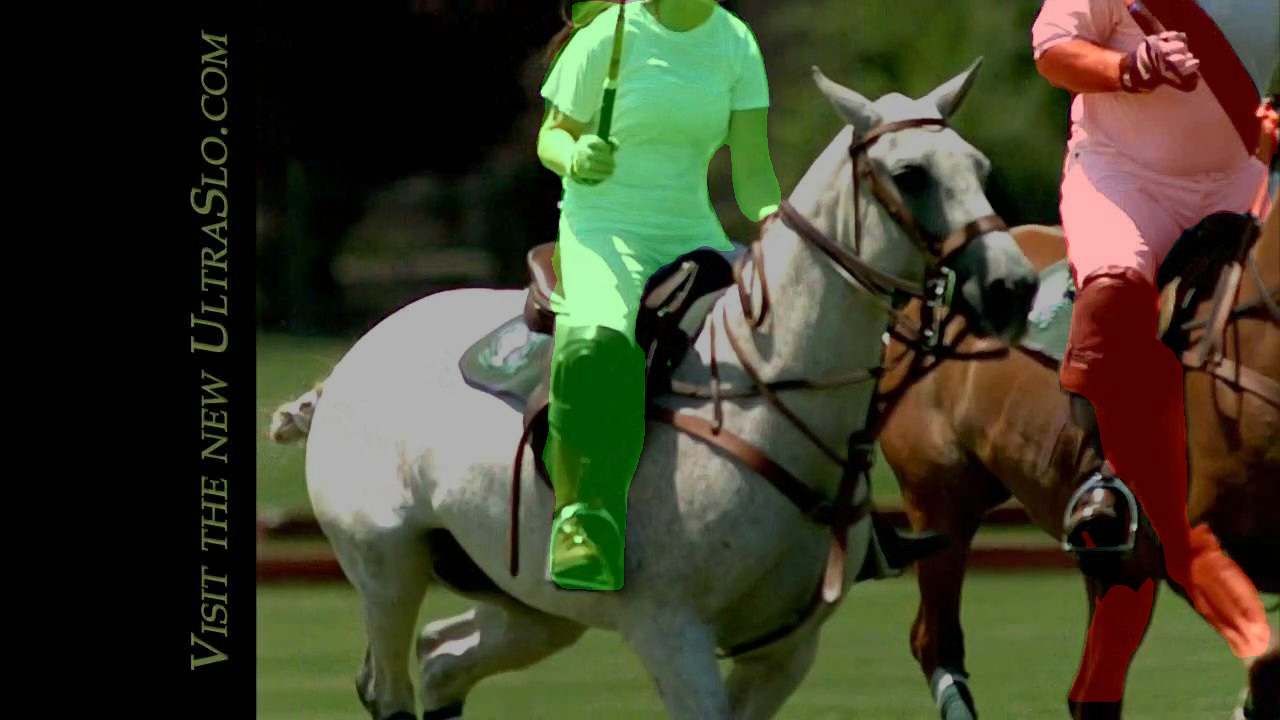}  & \hspace{-5mm}    
    \includegraphics[width=0.185\linewidth]{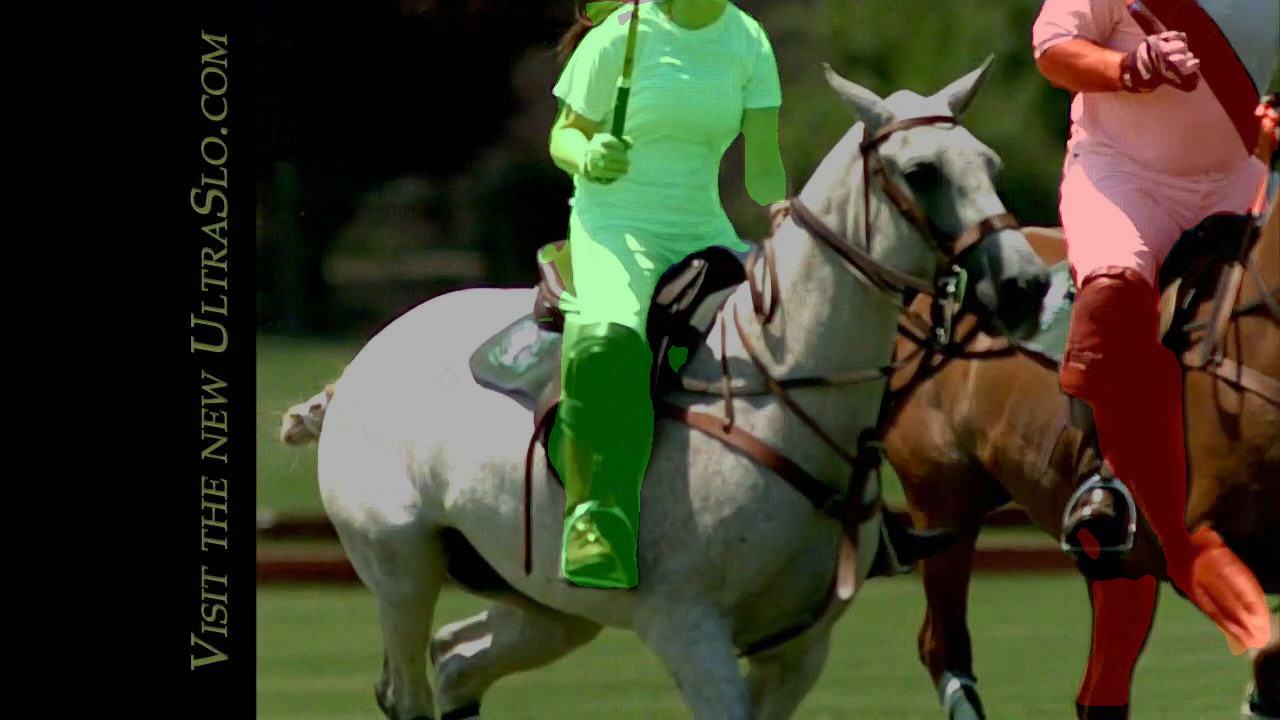} \\
    \specialrule{0mm}{-0.3mm}{-0.3mm}
    \hspace{-3.2mm}
    {\vspace{-1.5cm}}{(d)}& \hspace{-5mm}
    \includegraphics[width=0.185\linewidth]{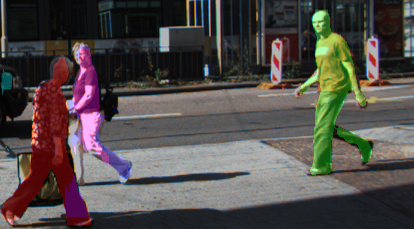}  & \hspace{-5mm}
    \includegraphics[width=0.185\linewidth]{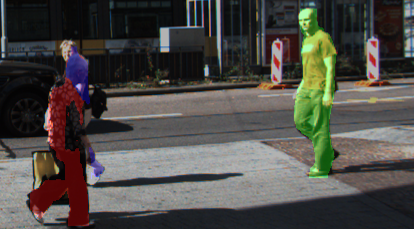}  & \hspace{-5mm}
    \includegraphics[width=0.185\linewidth]{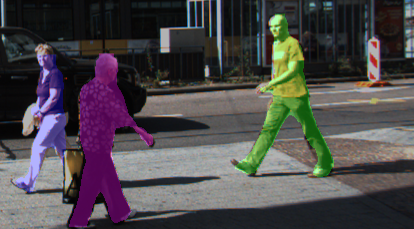}  & \hspace{-5.8mm}
    \includegraphics[width=0.185\linewidth]{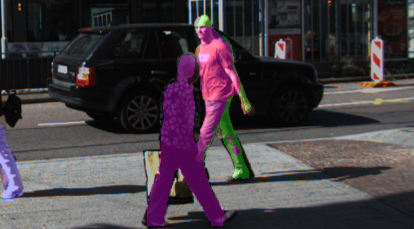}  & \hspace{-5.8mm}
    \includegraphics[width=0.185\linewidth]{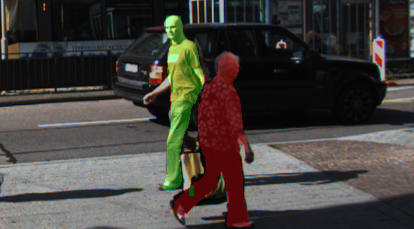} \\
    \specialrule{0mm}{-0.3mm}{-0.3mm}
    \hspace{-3.2mm}
    {\vspace{-1.5cm}}{(e)} & \hspace{-5mm}
    \includegraphics[width=0.185\linewidth]{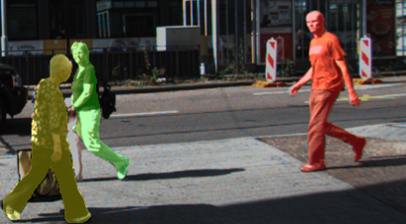}  & \hspace{-5mm}
    \includegraphics[width=0.185\linewidth]{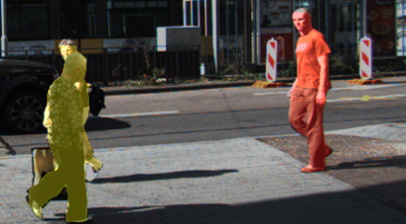}  & \hspace{-5mm}
    \includegraphics[width=0.185\linewidth]{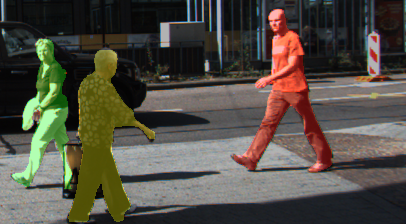}  & \hspace{-5mm}
    \includegraphics[width=0.185\linewidth]{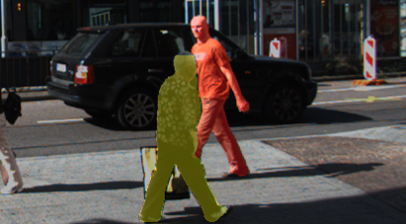}  & \hspace{-5mm}
    \includegraphics[width=0.185\linewidth]{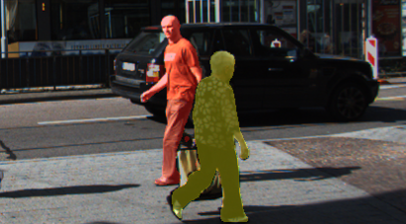} \\
    \specialrule{0mm}{-0.3mm}{-0.3mm}
    \hspace{-3.2mm}
    {\vspace{-1.5cm}}{(f)} & \hspace{-5mm}
    \includegraphics[width=0.185\linewidth]{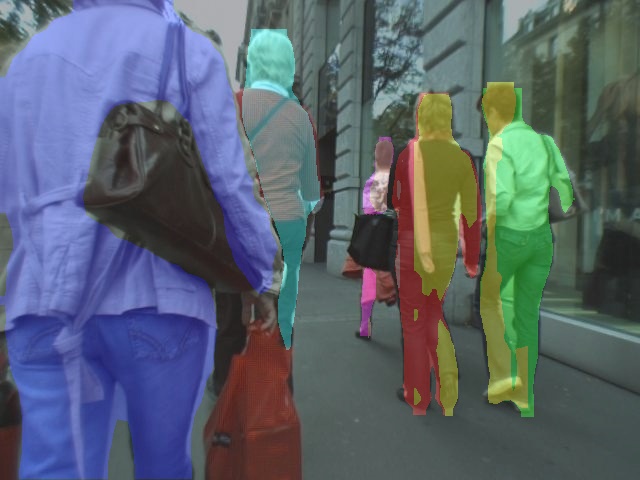}   & \hspace{-5mm}
    \includegraphics[width=0.185\linewidth]{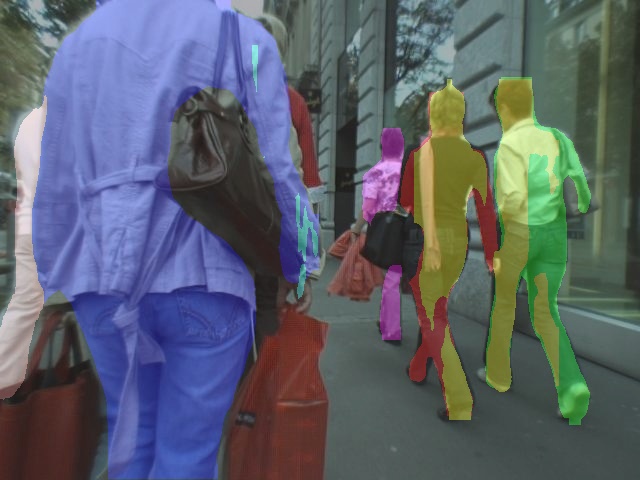}  & \hspace{-5mm}
    \includegraphics[width=0.185\linewidth]{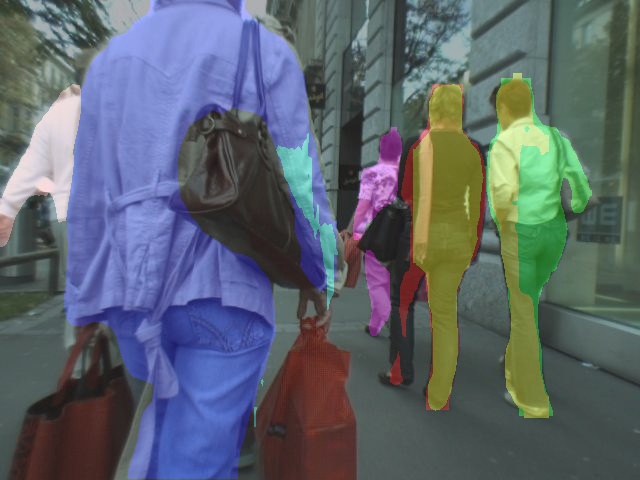}  & \hspace{-5mm}
    \includegraphics[width=0.185\linewidth]{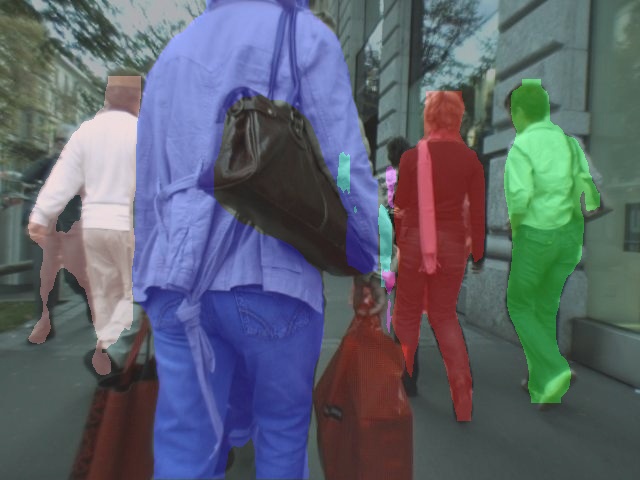}  & \hspace{-5mm}
    \includegraphics[width=0.185\linewidth]{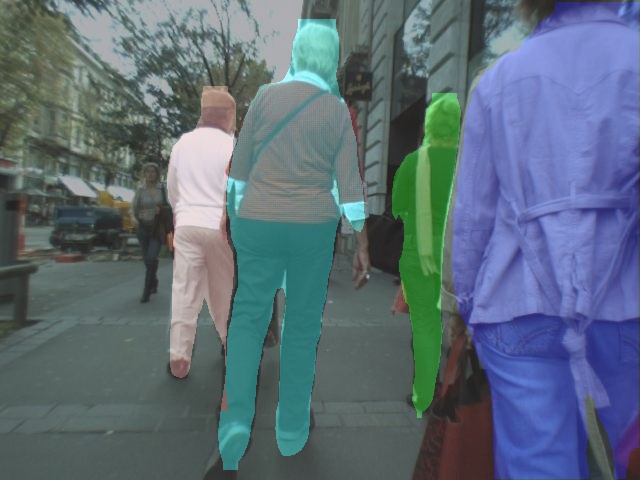} \\
    \specialrule{0mm}{-0.3mm}{-0.3mm}
    \hspace{-3.2mm}
     &\hspace{-5.8mm} \includegraphics[width=0.185\linewidth]{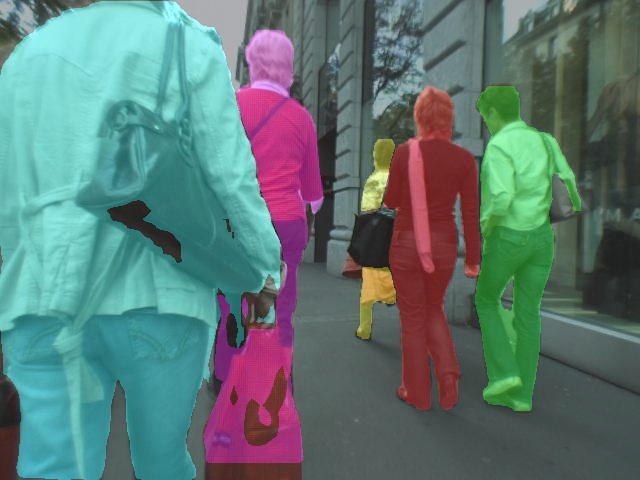}   & \hspace{-5mm}
    \includegraphics[width=0.185\linewidth]{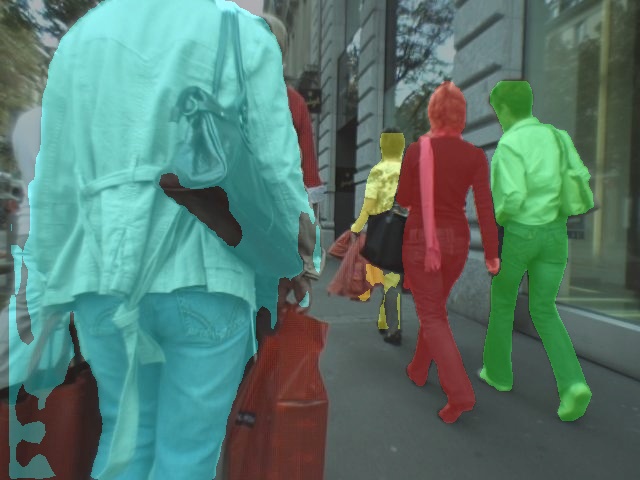}  & \hspace{-5mm}
    \includegraphics[width=0.185\linewidth]{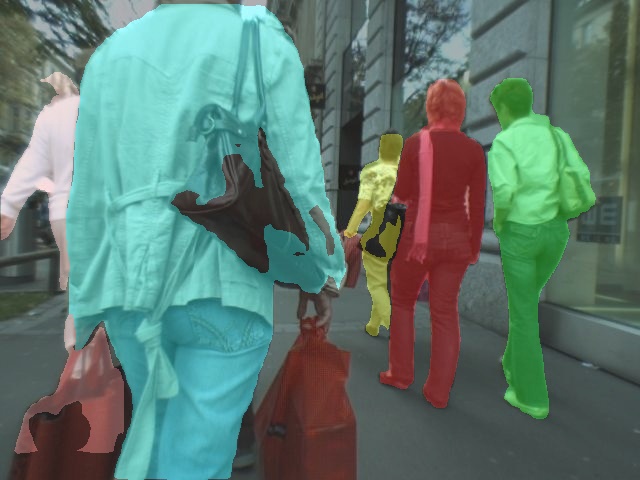}  & \hspace{-5mm}
    \includegraphics[width=0.185\linewidth]{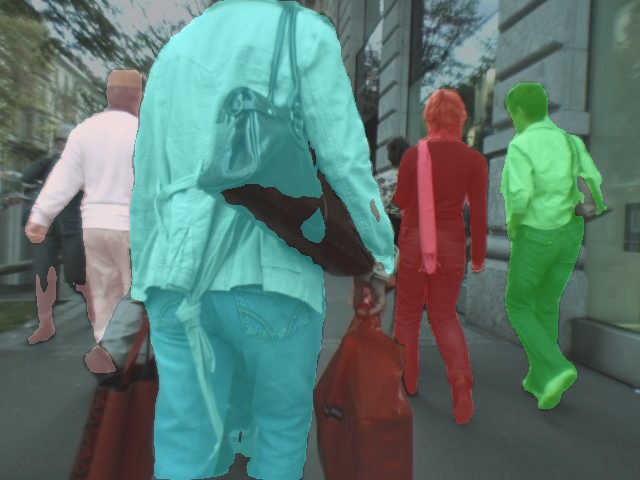}  & \hspace{-5mm}
    \includegraphics[width=0.185\linewidth]{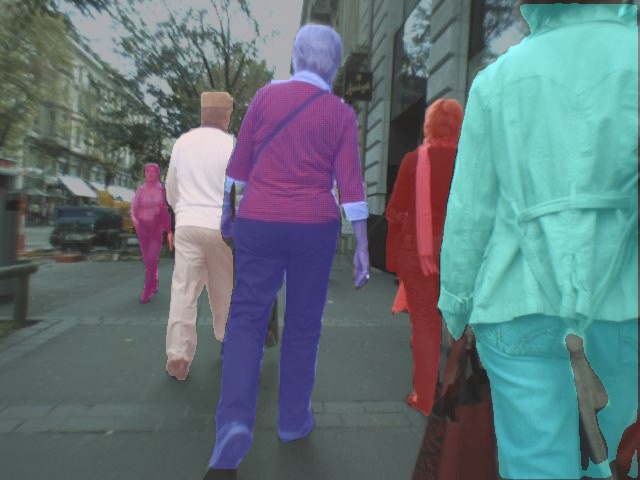} \\
    
  \end{tabular}
  \vspace{-2mm}
  \caption{Visual comparison of Masktrack and HVISNet. Each row demonstrates five frames sampled from a video sequence. (a), (c) and (e) show the experimental results of MaskTrack, while (b), (d), and (f) represent HVISNet results. The mask's color represents human identity, where the human in the same mask color has the same identity. }
  \label{fig:result}
\end{figure*}

We propose the dataset to evaluate the robustness of our method in complex scenes. 
We consider it better to utilize a public dataset for training and a dataset containing complex scenes for evaluation.
Therefore, during the dataset partition, the training set retains part of the person category in the VIS dataset. 
The missing part labels in the VIS indicate that some videos do not annotate all humans. 
We clean that part of the data and only keep the videos that match the above characteristics. In addition, we also add some other videos. The validation set is composed of 81 videos, including full complex scenes. 
The quantification of complex scenes is illustrated in detail in the next section.

\subsection{Dataset Statistic}
\label{subsec:data_statistic}

We illustrate the complexity of different datasets using three metrics: mask size of humans (\# small-target), number of instances (\# instances), and number of overlapping humans (\# overlapping). 
The size of the mask determines the difficulty of segmentation. 
Considering the obvious fact that the smaller the target, the greater the difficulty of segmentation, we calculate the mask of the human size for the distinction of target size. 
The human masks smaller than $32*32$ are marked as tiny, more extensive than $96*96$ are large, and the rest are medium. 
We count the proportion of small and medium-sized humans to measure the impact on the complexity of the dataset. 
The more the number of instances in the video, the more difficult it is to identify the identity, and the greater the difficulty. 
The number of overlapping instances in the video indicates the presence of occluded scenes in the dataset.

We also compared our validation set with the most similar dataset, the human portion of the VIS test dataset \cite{yang2019video} (VISPersons), as shown in Table~\ref{tab:dataset_compare}.
Despite the exhaustive annotations in VISPersons, its video scenes are relatively simple, and there are few videos of different instances of the same category or small annotations. Therefore, the VISPersons cannot evaluate the performance of the methods in complex video scenes such as occlusion and overlapping. 
Moreover, the complexity evaluation of our dataset is much higher than that of the VISPersons in terms of all these three aspects.

\section{Experiments}
\label{sec:exp}

We evaluate our proposed method and other baseline methods on two datasets: (1) HVIS, which is proposed in this paper and focuses on complex scenes with various kinds of occluded, overlapped, and deformed humans; (2) the person partition of Youtube-VIS~\cite{yang2019video}, which is proposed with the related task named VIS and contains relatively simple scenes. 

\PARbegin{Dataset.} 
The Youtube-VIS dataset~\cite{yang2019video} contains 40 common categories, including 2238 training videos, 302 validation videos, and 343 test videos. However, there are not any annotations in the released testing and validation sets. 
For comparison, we selected 647 videos with the category of humans from the training videos of Youtube-VIS and randomly split the subset of the Youtube-VIS dataset into 588 training videos and 59 test videos. 
For details of our proposed HVIS dataset, please refer to Section~\ref{sec:data} above.

\PARbegin{Evaluation Metrics.} 
Human-centric video segmentation in complex video scenes can be decomposed into detection, segmentation, and multi-object tracking. 
To evaluate our method from multiple task dimensions, we chose the evaluation metrics of related task MOTS~\cite{Voigtlaender19CVPR_MOTS}. 
The metrics contain soft multi-object tracking and segmentation accuracy (sMOTSA), multi-object tracking and segmentation accuracy (MOTSA), and mask-based multi-object tracking and segmentation precision (MOTSP). 
Among these metrics, sMOTSA considers detection, segmentation, and tracking quality simultaneously.

\PARbegin{Implementation Details.} 
We use a DLA-34~\cite{yu2018deep} backbone for HVISNet, and pre-train it on the COCO dataset~\cite{lin2014microsoft}. Our framework is implemented using Pytorch, which is trained end-to-end in 12 epochs with GeForce RTX 2080 Ti GPU. We set the initial learning rate to 0.01 and decreased the learning rate by a factor of 0.1 after 8 epochs and 11 epochs. 
Our model given a video frame with $512\times512$ resolution as inputs can run at about 30 FPS for inference.

\begin{table}[t] 
\centering
\caption{Performance of Center Inner Sampling on three instance segmentation methods (\%): BlendMask~\cite{chen2020blendmask}, CondInst~\cite{tian2020conditional}, and PolarMask~\cite{xie2020polarmask}. The models are trained on COCOPersons and tested on COCOPersons Val and OCHuman, respectively.}
\footnotesize
\setlength\tabcolsep{5pt}
\begin{tabular}{l|c|ccc|ccc}
\hline
\multicolumn{1}{c|}{\multirow{2}{*}{Method}} & \multicolumn{1}{c|}{\multirow{2}{*}{Strategy}} & \multicolumn{3}{c|}{COCOPersons} & \multicolumn{3}{c}{OCHuman} \\ \cline{3-8} 
\multicolumn{1}{c|}{} & \multicolumn{1}{c|}{} &$AP$ &$AP_M$ &$AP_H$ &$AP$ &$AP_M$ &$AM_H$ \\ \hline
&bbox &40.3 &44.4 &61.7 &25.7 &1.06    &27.0 \\
BlendMask &mask &41.0 &45.2 &62.6 &28.5 &\textbf{4.00} &29.8   \\
&ours &\textbf{41.2} &\textbf{45.6}  &\textbf{62.7} &\textbf{29.8} &2.10  &\textbf{31.1}\\ \hline
& bbox &39.8 &44.2 &61.4 &24.2 &2.00 &25.6 \\
CondInst &mask &40.1 &44.7 &62.0 & 27.5  &3.10 &28.9 \\
& ours & \textbf{40.1}&\textbf{44.9} &\textbf{62.0}&\textbf{28.1}   &\textbf{5.00}   &\textbf{29.3}\\ \hline
& bbox &34.3 &37.8 &51.8 &22.3 &3.90  &23.2  \\
PolarMask &mask &\textbf{34.6} &\textbf{38.1} &\textbf{52.5} &23.2 &3.90 &24.1   \\
&ours &34.5 &38.0 &52.2 &\textbf{23.4} &\textbf{4.40} &\textbf{24.2}\\ \hline
\end{tabular}

\label{tab:points}
\end{table}

\subsection{Main Results}
\label{subsec:main_results}

In this experiment, we compare the performance of our method with several state-of-the-art methods whose code is publicly available, such as MaskTrack~\cite{yang2019video} and STEm-Seg~\cite{athar2020stem}, on the HVIS dataset and VISPerson dataset.
Table~\ref{tab:result} presents the quantitative results. 
As shown, our method achieves competitive accuracy and speed under all evaluation metrics on HVIS and VISPerson. 
The main difference between our method and MaskTrack is that we build the framework based on a single-stage detector, simultaneously obtaining the mask and appearance embedding. 
Moreover, we use appearance embedding that represents humans' association to complete the association of human identities in the video, which can better distinguish different humans.
STEm-Seg introduces timing information through 3D convolution to better handle complex scenes, but the speed is low. 
The experimental results also illustrate this phenomenon. 
Our method can achieve similar accuracy at a real-time speed.

We visualize the several cases of MaskTrack and our method as shown in Figure~\ref{fig:result}. 
Each row demonstrates five frames sampled from a video sequence. (a), (c), and (e) show the experimental results of MaskTrack, while (b), (d), and (f) represent the results of HVISNet. 
As shown, when the posture of a person changes, MaskTrack treats it as a different individual. Therefore, MaskTrack cannot guarantee the identity consistency of the same person who is occluded by another human and reoccurs. 
Our method can mark humans accurately in these complex scenes. Even though the human size is small, our method also achieves good performance.

\subsection{Ablation study}
\label{subsec:ablation_study}

\begin{figure}[t]
\small
  \centering
  \includegraphics[width=0.48\textwidth]{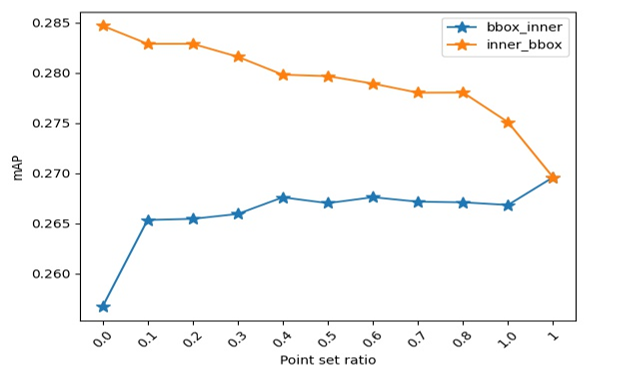}
  \caption{The impact of different positive samples on the instance segmentation method. }
  \label{fig:ablation}
  \vspace{-7pt}
\end{figure}

\begin{figure*}[th]
\centering
\begin{minipage}[t]{0.48\textwidth}
\centering
\includegraphics[width=0.98\linewidth]{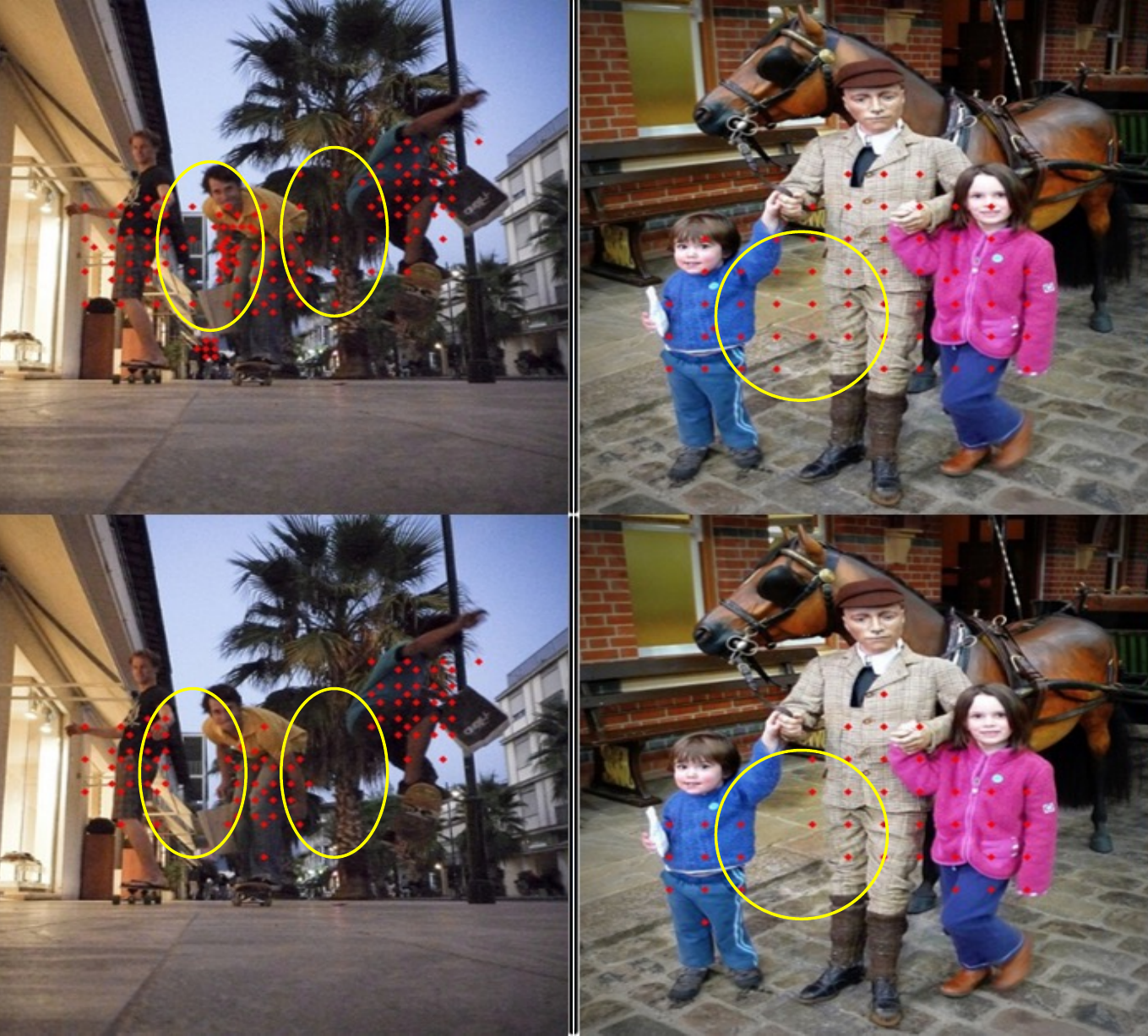}\\
\vspace{-5.0pt}\fontsize{7.0pt}{\baselineskip}\selectfont{(a)}
\end{minipage}
\begin{minipage}[t]{0.48\textwidth}
\centering
\includegraphics[width=0.98\linewidth]{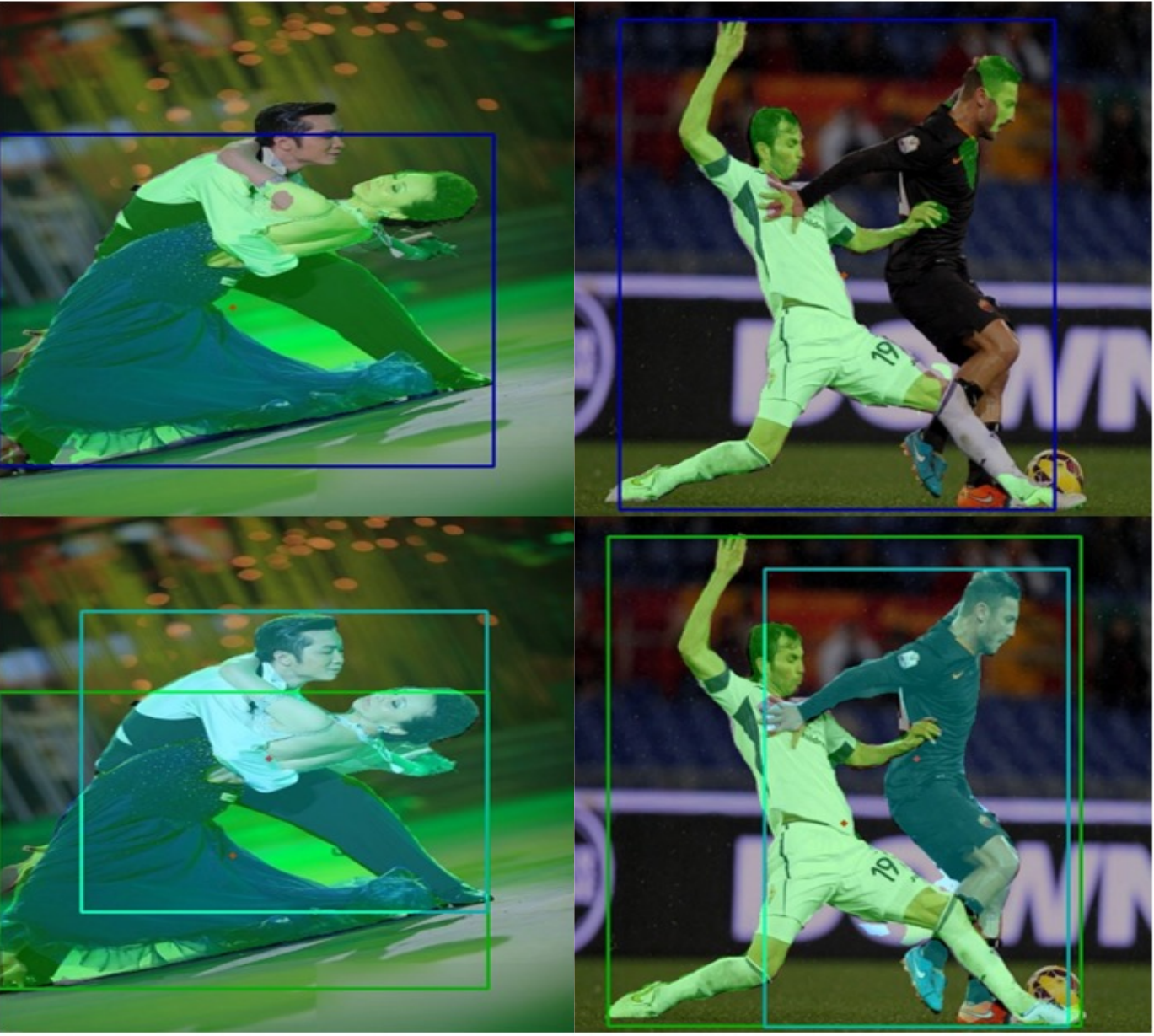}\\ 
\vspace{-5.0pt}\fontsize{7.0pt}{\baselineskip}\selectfont{(b)}
\end{minipage}
\caption{Qualitative diverse results of human image instance segmentation. These two rows of (a) and (b) represent the results of the sampling strategy in FCOS (upper) and our inner sampling strategy (lower), respectively. 
The red points in Figure~(a) indicate the obtained positions of the positive samples. Each red point corresponds to a bounding box and a score.
The yellow circles emphasize the difference between our strategy and FCOS.
Figure~(b) represents the segmentation and detection results of the sampling strategy in FCOS (upper) and our inner sampling strategy (lower), respectively.}
\label{fig:ablation_comp}
\end{figure*}

\PARbegin{Ambiguous Positive Sample Problem.} 
When we apply the sampling strategy proposed in FCOS to train the instance segmentation network, we discover the point features $ P_{NP}$ in complex scenes with poor performance. Although these features fed to the detection branch can lead to the bounding box's normal regression, the segmentation accuracy and appearance embedding differentiation are relatively low. The point features $ P_{NP}$ inside the bounding box but outside the mask are classified as positive samples in the classifiers in FCOS. Such a classifier can distinguish positive and negative samples in simple scenes such as single-human or multi-humans non-overlapping scenes. 
However, in complex scenes, such as severely occluded scenes, the positive samples of different instances overlap, with multiple labels during training. Thus, we explore whether different positive samples have an impact on the experimental results. We generate two positive sample sets $S_C$ and $S_I$ according to the strategy proposed in FCOS and center inner sampling, which proposes in this paper, respectively. Then, $a\% (a=10, 20, \cdots, 100)$ points in $S_I$ are augmented into $S_C$ and vice versa. And we use the two types of augmented positive samples to train the BlendMask. The model is trained on COCOPersons (the person category of COCO) with 90K iterations, tested on OCHuman \cite{zhang2019pose2seg}. The experimental results are shown in Figure~\ref{fig:ablation}. We can find that the different positive sample points impact the segmentation accuracy, and positive samples in the human masks are more conducive to segmentation.

\PARbegin{Inner Center Sampling on HIIS.} To verify the effectiveness of our proposed inner center sampling strategy, we evaluate the impact of three different sampling strategies on instance segmentation in complex scenes. 
The first strategy is the bounding-box center sampling proposed in FCOS, which results in ambiguous positive sample problems. 
The second strategy is centroid mask sampling associated with mask intuitively, which is more accurate. Moreover, it can alleviate ambiguous positive sample problems to a certain extent in some strange pose scenes, such as some humans dancing hip-hop. However, the performance is still deficient in the intense overlap scene. The third strategy is inner center sampling proposed in this paper, which alleviates ambiguous positive sample problems in complex scenes, especially for strong overlap. 
To illustrate the universality of the inner center sampling strategy, we apply it to other instance segmentation methods based on FCOS. The models are trained on COCOPersons (the person category of COCO) with 90K iterations and tested on COCOPersons and OCHuman datasets. From Table~\ref{tab:points}, the inner center sampling strategy leads to slight improvement on the COCOPersons dataset. Notably, the gain is much more significant on the OCHuman dataset, where each human instance is heavily occluded by one or several others. 
Figure~\ref{fig:ablation_comp} illustrates the qualitative comparison of human image instance segmentation between FCOS and our inner sampling strategy. The results show that our method effectively suppresses the direct interference samples of two targets.
Table~\ref{tab:comp_strategy} shows the comparison of these three sampling strategies. 
We count the positive sample numbers of three different strategies that fall inside the mask during the test. The values mean the percentage of positive samples (confident score \textgreater{0.5}) trained by the corresponding strategy.
Both Table~\ref{tab:points} and Table~\ref{tab:comp_strategy} demonstrate that our proposed strategy improves the overall performance by a significant margin, which shows the effectiveness of the plug-and-play inner center sampling strategy.

\begin{table}[t]
\caption{Quantitive comparison of three sampling strategies.}
\label{tab:comp_strategy}
\begin{tabular}{cccc}
\toprule
Confident Score &Center of bbox &Center of Mask &Inner-center \\ 
\midrule
\textgreater{}0.5 &0.891 &0.918     &0.931        \\ 
\bottomrule
\end{tabular}
\end{table}

\begin{table}[th]
\centering
\caption{Performance when constraining positive samples in segmentation (Seg.) and embedding (Emb.), respectively.}
\begin{tabular}{cc|ccc}
\toprule
Seg. &Emb. &sMOTSA &MOTSA  &MOTSP \\
\midrule
$\xmark$ &$\xmark$ &30.4 &40.3 &81.0 \\
$\checkmark$ &$\xmark$ &51.3 &63.9 &81.8 \\
$\xmark$ &$\checkmark$ &40.3 &51.8 &80.0 \\
$\checkmark$ &$\checkmark$ &\bf{52.1} &\bf{64.4} &\bf{81.9} \\
\bottomrule
\end{tabular}
\label{tab:hvisnet}
\end{table}

\PARbegin{Inner Center Sampling on Human-centric Video Segmentation.} 
We evaluate the impact of inner center sampling on the HVIS-CS in Table~\ref{tab:hvisnet}. We use the appearance embedding of a certain position on the feature map to represent the human's identity. Intuitively, it is more reasonable to use the embedding corresponding to the positive sample located inside the mask to represent the identity of the human. In the above experiment, the results show that the model can incorporate a plug-and-play inner center sampling strategy with a one-stage instance segmentation model to improve the model's segmentation accuracy. In the HVIS-CS task, we directly eliminate the ambiguous positive samples. In particular, we set the label of ambiguous positive samples to 0, which means treating ambiguous positive samples as negative samples. The result is shown in Table \ref{tab:hvisnet}, we find that if we directly eliminate the ambiguous positive samples, the performance of HVIS-CS is also improved. After we add the Inner Center Samping strategy on this basis, the performance achieves the best. Although our strategy can constrain the position of the positive samples within the mask as much as possible, it still cannot guarantee that all the positive sample points are inside the mask. Thus, we further constrain the positive samples by eliminating the ambiguous positive samples to improve the performance of the model.

\section{Conclusion}
\label{sec:conclusion}

In this paper, we propose a novel framework based on a one-stage detector for human-centric segmentation in complex video scenes, which had a good performance in accuracy and speed. Moreover, we found that the sampling strategies proposed in FCOS had a poor performance of segmentation and low accuracy of appearance embedding. We interpret this phenomenon as an ambiguous positive sample problem. To solve this problem, we proposed a novel inner center sampling strategy. 
Extensive experiments have been conducted to illustrate the effectiveness and universality of the inner center sampling strategy for performance improvement.
We also propose a benchmark HVIS better to evaluate the performance of different methods in complex scenes.
Comparative results on VISPersons and the proposed HVIS show that our framework achieves state-of-the-art performance.

\bibliographystyle{IEEEtran}
\bibliography{reference}

\end{document}